\documentclass{llncs} %

\usepackage{float}
\usepackage{graphicx}
\usepackage{wrapfig}

\usepackage[labelfont=bf]{caption}
\usepackage{verbatim}
\usepackage{color}
\usepackage{amsmath} 
\usepackage{mathtools} 
\usepackage{amsfonts} 
\usepackage{amssymb}
\usepackage[normalem]{ulem}

\usepackage{array}
\usepackage{booktabs}
\usepackage{rotating}
\usepackage{multirow}
\usepackage{multicol}
\usepackage{lscape}
\usepackage{subfig}
\usepackage{comment}

\usepackage{wrapfig}
\usepackage{rotating}
\usepackage{epstopdf}

\usepackage[export]{adjustbox}

\usepackage{algorithm,algorithmicx,algpseudocode}

\usepackage{lineno}
\usepackage{soul}
\usepackage{color}
\usepackage{xcolor}
\usepackage{xargs}

\definecolor{brilliantrose}{rgb}{1.0, 0.33, 0.64}
\definecolor{capri}{rgb}{0.0, 0.75, 1.0}
	
\usepackage[colorlinks=true,linkcolor=capri,citecolor=capri]{hyperref}

\usepackage[colorinlistoftodos,prependcaption,textsize=tiny]{todonotes}
\newcommandx{\ISLEM}[2][1=]{\todo[linecolor=brilliantrose,backgroundcolor=brilliantrose!25,bordercolor=brilliantrose,#1]{#2}}

\usepackage{tikz,xcolor,hyperref}

\definecolor{lime}{HTML}{A6CE39}
\DeclareRobustCommand{\orcidicon}{
	\begin{tikzpicture}
	\draw[lime, fill=lime] (0,0) 
	circle [radius=0.16] 
	node[white] {{\fontfamily{qag}\selectfont \tiny ID}};
	\draw[white, fill=white] (-0.0625,0.095) 
	circle [radius=0.007];
	\end{tikzpicture}
	\hspace{-2mm}
}

\foreach \x in {A, ..., Z}{\expandafter\xdef\csname orcid\x\endcsname{\noexpand\href{https://orcid.org/\csname orcidauthor\x\endcsname}
			{\noexpand\orcidicon}}
}
			
\usepackage{color}

\definecolor{darkgreen}{rgb}{0.53, 0.66, 0.42}

\begin{document}

\title{Residual Embedding Similarity-Based Network Selection for Predicting Brain Network Evolution Trajectory from a Single Observation}

\titlerunning{Short Title}  

\author{Ahmet Serkan G\"{o}kta\c{s}\index{G\"{o}kta\c{s}, Ahmet Serkan}\inst{1}, Alaa Bessadok\orcidB{}\inst{1,2} \and Islem Rekik\orcidA{}\inst{1} \index{Rekik, Islem}\thanks{ {corresponding author: irekik@itu.edu.tr, \url{http://basira-lab.com}} This work is accepted for publication in the PRedictive Intelligence in MEdicine (PRIME) workshop Springer proceedings in conjunction with MICCAI 2020.}}

\institute{$^{1}$ BASIRA Lab, Faculty of Computer and Informatics, Istanbul Technical University, Istanbul, Turkey \\ $^{2}$ LATIS Lab, ISITCOM, University of Sousse, Sousse, Tunisia}

\authorrunning{S. G''{o}kta\c{s} et al.}

\maketitle              

\begin{abstract}

Predicting the evolution trajectories of brain data from \emph{a baseline timepoint} is a challenging task in the fields of neuroscience and neuro-disorders. While existing predictive frameworks are able to handle Euclidean structured data (i.e, brain images), they might fail to generalize to \emph{geometric non-Euclidean} data such as brain networks. Recently, a seminal brain network evolution prediction framework was introduced capitalizing on learning how to select the most similar training network samples at baseline to a given testing baseline network for the target prediction task. However, this rooted the sample selection step in using Euclidean or learned similarity measure between \emph{vectorized} training and testing brain networks. Such sample connectomic representation might include irrelevant and redundant features that could mislead the training sample selection step. Undoubtedly, this  fails to exploit and preserve the topology of the brain connectome. To overcome this major drawback, we propose Residual Embedding Similarity-Based Network selection (RESNets) for predicting brain network evolution trajectory from a single timepoint. RESNets first learns a compact geometric embedding of each training and testing sample using adversarial connectome embedding network. This nicely reduces the high-dimensionality of brain networks while preserving their topological properties via graph convolutional networks. Next, to compute the similarity between subjects, we introduce the concept of a connectional brain template (CBT), a fixed network reference, where we further represent each training and testing network as a deviation from the reference CBT in the embedding space. As such, we select the most similar training subjects to the testing subject at baseline by comparing their learned residual embeddings with respect to the pre-defined CBT. Once the best training samples are selected at baseline, we simply average their corresponding brain networks at follow-up timepoints to predict the evolution trajectory of the testing network. Our experiments on both healthy and disordered brain networks demonstrate the success of our proposed method in comparison to RESNets ablated versions and traditional approaches.  Our RESNets code is available at \url{http://github.com/basiralab/RESNets}.

\end{abstract}


\section{Introduction}

Longitudinal neuroimaging of the brain has spanned several neuroscientific works to  examine early disease progression and eventually improve neurological disorder diagnosis\cite{Yang:2019,Zhou:2019}. Existing studies aiming to predict brain evolution trajectories from a single baseline timepoint are mainly focused on Euclidean neuroimaging data such as magnetic resonance imaging (MRI). For instance, \cite{Rekik:2017a} predicted the multishape trajectory of the baby brain using neonatal MRI data. Similarly, \cite{Gafuroglu:2018} and \cite{consistent_aging} used MR images to predict brain image evolution trajectories for early dementia detection. Although pioneering, such works mainly focused on Euclidean structured data (i.e, images), which is a flat representation of the brain and does not reflect the connectivity patterns existing among brain regions encoded in brain networks (i.e, connectomes). Specifically, a brain network is a graph representation of interactions in the brain between a set of anatomical regions of interests (ROIs) (or nodes). Such interactions are encoded in the edge weights between pairs of ROIs, capturing the function, structure, or morphology of the brain as a complex highly interconnected system.

\begin{figure}[h]
\begin{center}
\includegraphics[width=12cm]{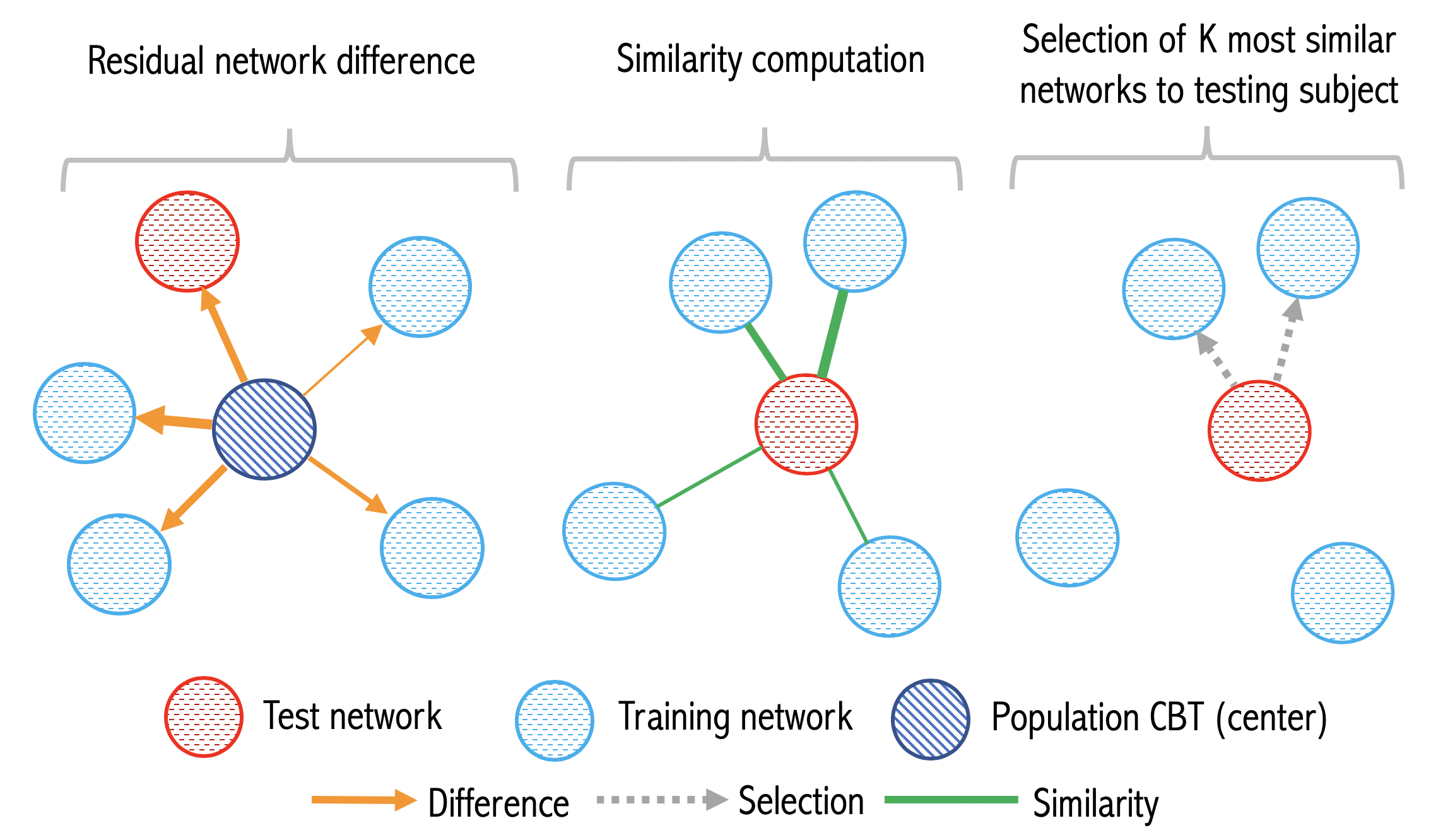}
\end{center}
\caption{ \emph{Selection principle of similar training subjects to a testing subject at a baseline timepoint $t_0$.} In this illustration, we sketch the scenario where we first generate the residual networks by computing an element-wise absolute difference between each network and population-driven connectional brain template (CBT). Second, we calculate the similarities between subjects using the resulting residual networks. Finally, we select top $K$ similar subjects in the population to the testing one at baseline for the target prediction task.} 
\label{fig:0} 
\end{figure}

So far, we have identified a single work on brain network evolution trajectory prediction \cite{baha}, leveraging multi-kernel manifold learning technique to predict follow-up brain networks from a baseline network. This landmark work predicted the spatiotemporal trajectory of a network by first selecting the closest neighboring training samples (i.e., brain networks) to the testing sample at baseline via learning their pairwise similarities. Second, by averaging the follow-up selected training networks at later timepoints, the evolution trajectory of a testing brain network was generated.  However, such approach is limited by the vectorization of baseline brain networks to learn the similarities between pairs of brain networks. Clearly, this fails to preserve and exploit both local and global topologies of the brain connectome \cite{Fornito:2015}. In fact, each brain region has a particular topological property underpinning its function and which can be changed with healthy or atypical aging. The vectorization of brain networks to extract the sample features (i.e., connectivity weights) is widely adopted in connectomic machine learning tasks such as classification \cite{Wang:2020,Richiardi:2010}. However, such connectome feature representations spoil the rich topological properties of the brain as a graph including its percolation threshold, hubness and modularity \cite{Bassett:2017}. A second limitation of \cite{baha} lies in comparing pairs of brain networks at baseline without considering their inherently shared connectivity patterns. In other words, one can think of each individual sample as a deviation from a training population center capturing the shared traits across training samples.  Hence, estimating a fixed population network, namely a connectional brain template (CBT) as introduced in \cite{cbt}, presents a powerful tool to integrate complementary information from different brain networks. To capture shared connectivity patterns between training and testing samples, we propose to define a `normalization' or `standardization' process of brain networks (\textbf{Fig.~\ref{fig:0})}. Eventually, we hypothesize that reducing inter-subject variability in the baseline training network population through a normalization process will contribute towards helping better identify the best neighboring samples to the testing sample by modeling them as deviations from the `standard/normalized' brain network representation (CBT). 

This recalls an analogous brain imaging protocol, where one registers an image to an atlas image for group comparison and building classification models \cite{Liu:2015} following the extraction of `shared' features in the standard shared space. Following this lead, we will use the estimated CBT to (1) first \emph{normalize} each individual baseline brain connectome to the estimated CBT at baseline timepoint by generating the residual between each connectome and the fixed CBT, (2) then use the normalized individual network (i.e. residual with respect to the population CBT) at baseline to guide the prediction of follow-up brain connectomes acquired at later timepoints. To do so, we leverage netNorm \cite{cbt} which produces a unified normalized connectional representation of a population of brain networks. Notably, this proposed strategy only solves the second limitation, while the first drawback of existing works remains untackled. 

To overcome both limitations in a unified framework, we propose Residual Embedding Similarity-based Network selection (RESNets) for predicting brain connectome evolution trajectory from a single timepoint. To learn a topology-preserving brain network representation in a low-dimensional space for the target sample selection task, we first propose to learn the embedding of each baseline network by leveraging adversarial connectome embedding (ACE) \cite{ace}. ACE is rooted in the nascent field of geometric deep learning \cite{Bronstein:2017,Hamilton:2017} where a brain graph is auto-encoded  using an encoder $E$, defined as a Graph Convolution Network (GCN) \cite{Kipf:2016}, and regularized by a discriminator $D$ aiming to align the distribution of the learned connectomic embedding with that of the original connectome. Specifically, we use ACE to embed each training and testing brain network as well as the CBT in a fully independent manner. Next we define the residual embedding of each training and testing sample using the absolute difference between the sample embedding and the CBT embedding. The resulting residual embeddings represent how each brain network deviate from the training population center. To predict the follow-up brain networks of a testing subject, we first compute the cosine similarities between training and testing subjects using their CBT-based residual embeddings, then identify the closest residual embeddings to the testing subject at baseline. Finally, we average their corresponding training networks at consecutive timepoints to ultimately predict the brain network evolution trajectory of a testing subject.

\begin{sidewaysfigure}
\centering
\includegraphics[width=19cm]{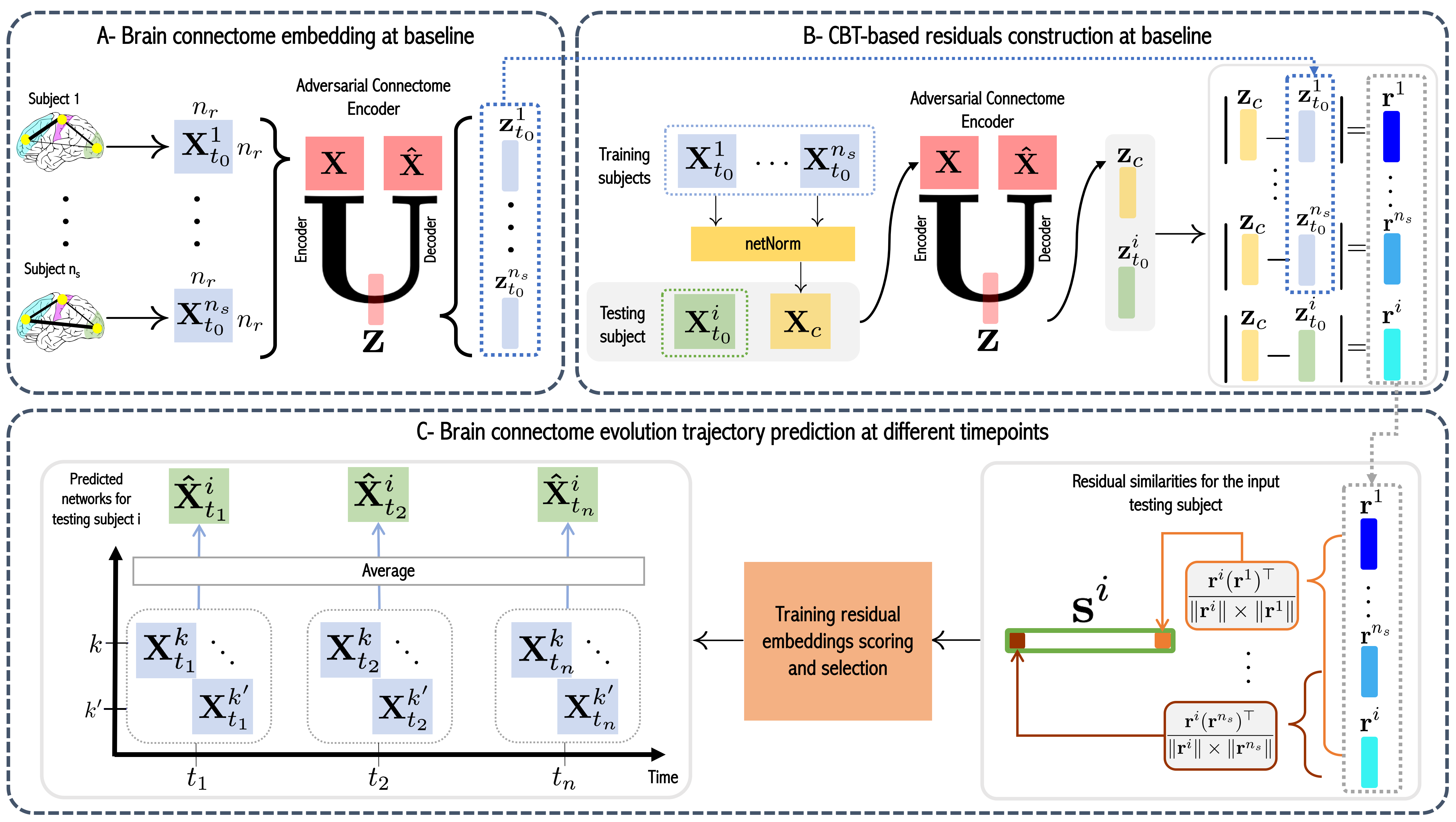}
\caption{ \emph{Proposed frameworks steps for predicting brain network evolution trajectory from a single observation at baseline timepoint $t_0$ using residual embeddings similarities.} \textbf{(A) Adversarial brain network embedding at baseline $t_{0}$.} We learn the embeddings of each training brain network in the population at baseline using an adversarial connectome embedding encoder. \textbf{(B) CBT-based residuals construction.} We first estimate the CBT of the training networks and learn the embedding of both generated CBT and the testing brain network. Next, we normalize each brain network by computing the residual between the networks and the CBT embeddings. \textbf{(C) Brain network prediction at different timepoints.}
To predict a follow-up brain network of a testing subject, we (i) compute the cosine similarities between subjects using their CBT-based residual embeddings, (ii) select the top similar subjects to the testing one and (iii) average their corresponding brain networks at a follow-up timepoint $t_{h}$ where $\ h \in \{1, \dots, n_{t}\}$.}
\label{fig:1}
\end{sidewaysfigure} 

\section{Proposed Method}

\textbf{Problem Definition.} A brain network can be represented as \( \mathbf{N}=\{ \mathbf{V},\mathbf{E},\mathbf{X} \}\) where \(\mathbf{V}\) is a set of nodes (i.e, ROIs) and \(\mathbf{E}\) is a set of weighted edges encoding the interaction (connectivity) between each pair of nodes. Let \(\mathbf{X} \in \mathbb{R}^{n_{r} \times n_{r}} \) denote the connectivity matrix where \({n_{r}}\) is the number of nodes (ROIs) in a connectome. Each training sample \(s\) in our dataset is represented by a set of \emph{time-depending} brain connectomes $\{ \mathbf{X}_{t_{g}}^s \}_{g=0}^{n_{t}}$, each measured at a particular timepoint $t_{g}$. Given a testing connectome $i$ solely represented by a brain network at first timepoint $t_{0}$ denoted as $\mathbf{X}_{t_{0}}^{i}$, our objective is to predict its missing brain networks $\{ \mathbf{\hat{X}}^i_{t_h} \}$ at later timepoints $t_{h}, \ h \in \{1,...,n_{t}\}$. 

In the following, we present the main steps of our network evolution trajectory prediction framework from a baseline observation. \textbf{Fig.}~\ref{fig:1} provides an overview of the key
three steps of the proposed framework: 1) adversarial connectome embedding of training and testing brain networks and the CBT estimated by netNorm at baseline, 2) construction of residual embeddings, and 3) prediction of brain networks at follow-up timepoints. For easy reference, we summarize the major mathematical notations in \textbf{Table}~\ref{1}.

\begin{table}
\caption{\emph{Major mathematical notations used in this paper}.\label{1}}
\centering
\begin{scriptsize}
\begin{tabular}{ >{\centering\arraybackslash}m{1in}  >{\centering\arraybackslash}m{4in} }
	\toprule
	Mathematical notation & Definition \\
	\midrule
	$n_t$ & number of timepoints (baseline and follow-up) \\
	
	$n_{s}$ & number of training subjects \\
	
	$n_{r}$ & number of region of interest (ROI) in a brain network \\
	
	$d$ & number of features of the embedded graph \\
	
	$K$ & number of neighbors of a testing subject \\
	
	$\mathbf{X}_{t_{0}}^{i}$ & brain network of a testing subject $i$ at the baseline timepoint ${t_{0}}$ in $\mathbb{R}^{n_{r} \times n_{r}}$\\
	
	$\mathbf{X}_{t_{0}}^{j}$ & brain network of a training subject $j$ at the baseline timepoint ${t_{0}}$ where $j\in{\{1,...,n_{s}\}}$ in $\mathbb{R}^{n_{r} \times n_{r}}$ \\
	
	$\mathbf{\hat{X}}_{t_{h}}^{i}$ & predicted brain network of a testing subject $i$ at a follow-up timepoint $t_{h}$ where $h\in{\{1,...,n_{t}\}}$\\
	
	$\mathbf{X}_{t_{h}}^{l}$ & brain network of a training neighbor $l\in{\{k,...,k'\}}$ to the testing subject at a follow-up timepoint $t_{h}$ in $\mathbb{R}^{n_{r} \times n_{r}}$ \\
	
	$\mathbf{X}_{c}$ & connectional brain template (CBT) of the training brain networks in $\mathbb{R}^{n_{r} \times n_{r}}$ \\
	
	$\mathbf{s}^{i}$ & similarity vector between the testing $i$ and training subjects $n_{s}$ using their CBT-based residuals in $\mathbb{R}^{1 \times n_{s}}$ \\
	
	$\mathbf{z}_{c}$ & learned CBT embedding of the training population in $\mathbb{R}^{d \times 1}$\\
	
	$\mathbf{z}_{t_{0}}^{i}$ & learned brain network embedding of a testing subject $i$ at the baseline timepoint ${t_{0}}$ in $\mathbb{R}^{d \times 1}$\\
	
	$\mathbf{z}_{t_{0}}^{j}$ & learned brain network embedding of a training subject $j$ in $\mathbb{R}^{d \times 1}$\\
	
	$\mathbf{r}^{i}$ &  testing CBT-based residual in $\mathbb{R}^{d \times 1}$\\
	
	$\mathbf{r}^{j}$ &  training CBT-based residual in $\mathbb{R}^{d \times 1}$\\
	
	$\mathbf{v}_{ab}^{j}$ & similarity value for subject j related to ROIs $a$ and $b$\\
	
    $\mathbf{H}_{ab}$ & high-order graph (graph of a graph) for pair of ROIs $a$ and $b$ $\in \mathbb{R}^{n_s \times n_s}$\\
    
    $\mathbf{D}_{ab}(j)$ & cumulative distance of node $j$ in $\mathbf{H}_{ab}$\\ 
    
    $E(\mathbf{z}_{t_{0}} \vert \mathbf{F}_{t_{0}},\mathbf{X}_{t_{0}})$ & encoder used for learning the brain network embedding at a baseline timepoint taking as input a feature matrix $\mathbf{F}_{t_{0}}$ and a brain network $\mathbf{X}_{t_{0}}$ \\
    
    $D(\mathbf{\hat{X}}_{t_{0}} \vert \mathbf{z}_{t_{0}})$ & decoder used to reconstruct the brain network from its embedding $\mathbf{z}_{t_{0}}$\\
    
    $\mathbf{\mathcal{D}}$ & discriminator used for connectome embedding regularization taking as input the real brain network $ \mathbf{X}_{t_{0}}^s $ and the embedded network $\mathbf{z}_{t_{0}}^s$ of a subject $s$\\ 
\bottomrule
\end{tabular}  
\end{scriptsize}
\end{table}

\textbf{A- Adversarial brain network embedding at baseline $t_0$.} We propose in this step to learn a low-dimensional topology-preserving representation of a given brain network at baseline using ACE model \cite{ace}, which is a subject-based adversarial embedding autoencoder tailored for brain connectomes (\textbf{Fig.}~\ref{fig:1}-A). ACE architecture is composed of a graph convolution network (GCN) \cite{Kipf:2016} encoder $E(\mathbf{z} \vert \mathbf{F},\mathbf{X})$ with two layers inputting a feature matrix $\mathbf{F}$ and an adjacency matrix $\mathbf{X}$. Since nodes in a brain connectome have no features, we filled $\mathbf{F}$ matrix with identify values (a set of '1'). At baseline timepoint $t_0$, we define the layers of our encoder $E(\mathbf{z}_{t_{0}} \vert \mathbf{F}_{t_{0}},\mathbf{X}_{t_{0}})$ and the graph convolution function used in each mapping layer as follows:

\begin{gather}
	 \mathbf{z}_{t_{0}}^{(l)} = f_{\phi}(\mathbf{F}_{t_{0}}, \mathbf{X}_{t_{0}} \vert \mathbf{W}^{(l)}); 
    	\quad\text{ }
    	{f}_{\phi}(\mathbf{F}_{t_{0}}^{(l)}, \mathbf{X}_{t_{0}} \vert \mathbf{W}^{(l)}) = {\phi}(\mathbf{\widetilde{D}}^{-\frac{1}{2}}\mathbf{\widetilde{\mathbf{X}}}_{t_{0}}\mathbf{\widetilde{D}}^{-\frac{1}{2}}\mathbf{F}_{t_{0}}^{(l)}\mathbf{W}^{(l)}),
\label{eq:1}
\end{gather}

$\mathbf{z}_{t_{0}}^{(l)}$ represents the resulting brain network embedding of the layer $l$. $\phi$ is the Rectified Linear Unit (ReLU) and linear activation functions we used in the first and second layers, respectively. $\mathbf{X}_{t_{0}}$ denotes the input brain network connectivity matrix at baseline timepoint. $\mathbf{W}^{(l)}$ is a learned filter encoding the graph convolutional weights in layer $l$. $f(.)$ is the graph convolution function where $\mathbf{\widetilde{\mathbf{X}}}_{t_{0}} = \mathbf{\mathbf{X}}_{t_{0}} + \mathbf{I}$ with $\mathbf{I}$ is the identity matrix used for regularization, and $\mathbf{\widetilde{D}}_{aa} = \sum_{b}\mathbf{\widetilde{\mathbf{X}}}_{t_{0}}(ab)$ is a diagonal matrix storing the topological strength of each node. We note that ACE is trained for each sample independently to learn its embedding. The \emph{individual-based} learning of brain network embedding yields not only to reducing the high-dimensionality of the original brain network but also preserving its topology via a set of layer-wise graph convolutions. To decode the resulting connectomic embedding $\mathbf{z}_{t_{0}}$, we compute the sigmoid function of the embedding $\mathbf{z}_{t_{0}}(a)$ and the transposed embedding $\mathbf{z}_{t_{0}}(b)$ of nodes $a$ and $b$, respectively. Hence, we define our decoder $D(\mathbf{\hat{X}}_{t_{0}} \vert \mathbf{z}_{t_{0}})$ and the reconstruction error $\mathbf{\mathcal{L}}$ as follows: 

\begin{gather}
	{D}(\mathbf{\hat{X}}_{t_{0}} \vert \mathbf{z}_{t_{0}}) = \frac{\mathrm{1} }{\mathrm{1} + e^{-(\mathbf{z}_{t_{0}}(a) \cdot \mathbf{z}_{t_{0}}^\top(b))} };
	\quad\text{ }
	\mathbf{\mathcal{L}} = \mathbf{E}_{E(\mathbf{z}_{t_{0}} \vert \mathbf{F}_{t_{0}},\mathbf{X}_{t_{0}})} [ \log{D}(\mathbf{\hat{X}}_{t_{0}} \vert \mathbf{z}_{t_{0}}) ]
\end{gather}

Moreover, each brain network embedding is adversarially regularized using a discriminator $\mathbf{\mathcal{D}}$ that aligns the distribution of learned embedding $\mathbf{z}_{t_{0}}^{(l)}$ in the last encoding layer $l$ towards the prior data distribution that is the real baseline brain network $\mathbf{X}_{t_{0}}$. In particular, $\mathbf{\mathcal{D}}$ is a multilayer perceptron aiming to minimize the error in distinguishing between real and fake data distributions. We formulate the adversarial brain network embedding cost function at a first timepoint $t_{0}$ as follows:

\begin{gather}
	\min_{E} \max_{\mathbf{\mathcal{D}}} \mathbf{E}_{p_{(real)}}[\log\mathbf{\mathcal{D}}(\mathbf{X}_{t_{0}})] + \mathbf{E}_{p_{(fake)}}[\log{(1 - \mathbf{\mathcal{D}}(\mathbf{z}_{t_{0}}^{(l)})))}]
	\label{eq:2}
\end{gather}

where $\mathbf{E}$ is the cross-entropy cost. $E$ and $\mathbf{\mathcal{D}}$ represent our GCN encoder and discriminator, respectively.

\textbf{B- CBT-based residual construction.} To compute the similarity between training and testing brain networks, we propose to consider the inherently shared connectivity patterns which are captured in an `average' population network called connectional brain template (CBT). To this aim, we leverage netNorm \cite{cbt} which estimates a normalized connectional map of a population of brain networks (\textbf{Fig.}~\ref{fig:1}-B). Specifically, we learn a CBT for the training baseline brain connectomes in four consecutive stages. Since netNorm was originally designed to handle multi-view brain networks, where each subject is represented by a set of multimodal networks, we adapt it to our aim of integrating a set of \emph{uni-modal} brain networks. Firstly, for each subject, we extract the value of the similarity between ROIs $a$ and $b$ as follows:

\begin{gather}
	\mathbf{v}^{j}_{ab} = \mathbf{X}^{j}(a, b)\:;\forall \:\:1\leq j\leq n_s
\end{gather}

where $\mathbf{X}^j$ represents the brain network of the subject $j$. Secondly, using these extracted values, we construct the high-order graph, storing for each pair of subjects $j$ and $j'$, the Euclidean distance between their corresponding connectivity weights between ROIs $a$ and $b$ as follows:

\begin{gather}
	\mathbf{H}_{ab}(j, j{}') = \sqrt{(\mathbf{v}_{ab}^{j}- \mathbf{v}_{ab}^{j{}'})^{2} }\:;\forall \:\:1\leq j, j{}'\leq n_s\end{gather}

This high-order graph will constitute the basis of selecting the connectivity weight between ROIs $a$ and $b$ of the most centered subject $(j)$ with respect to all other subjects (i.e., achieving the lowest distance to all samples). To do so, we use cumulative distance metric for each subject $j$ as:

\begin{gather}
	\mathbf{D}_{ab}(j)=\sum_{j{}' = 1}^{n_s}\mathbf{H}_{ab}(j, j{}') = \sum_{j{}' = 1}^{n_s}\sqrt{(\mathbf{v}_{ab}^{j}- \mathbf{v}_{ab}^{j{}'})^{2} }\:;\forall \:\:1\leq j, j{}'\leq n_s
\end{gather}

Notably, this defines the strength of node $j$ in the high-order graph $H_{ab}$. In the last step, we define the connectivity weight in the centered final CBT denoted by $\mathbf{X}_c$ as follows:
\begin{gather}
	\mathbf{X_c}(a, b) = \mathbf{v}_{ab}^{j'}; \: where \: j'=\min_{1\leq j \leq n_s}D_{ab}(j)
\end{gather}

Originally, to fuse these matrices into a single connectome, we need to use a network fusion method \cite{Wang:2014} which reduces the tensor into a single representative matrix. However, since we design our framework for a single-view brain network evolution prediction, we skip the network fusion step of the multi-view CBT estimation.

Next, we feed the resulting CBT denoted by $\mathbf{X}_c$ to the GCN encoder $E$ to learn its embedding $\mathbf{z}_c$ using \textbf{Eq.}~\ref{eq:1}. Last, we compute the residual embeddings using the following formula: $\mathbf{r} = | \mathbf{z}_{c} - \mathbf{z}_{t_{0}} |$, where $\mathbf{z}_{t_{0}}$ is the network embedding of a subject in the population (\textbf{Fig.}~\ref{fig:1}-B). By producing these residuals, we are normalizing each baseline brain connectome to a fixed brain network reference (i.e, CBT) of the whole population.

\textbf{C- Brain network prediction at different timepoints.} To predict the evolution trajectory of a testing brain network, we first search its most similar training networks at baseline timepoint $t_{0}$ then average their corresponding brain networks at later timepoints $\{ t_{h} \}_{h=1}^{n_{t}}$. To this end, we propose to select subjects based on their learned residual embeddings (\textbf{Fig.}~\ref{fig:1}-C). Specifically, we project the testing subject residual $\mathbf{r}^i$ on each training subject $\mathbf{r}^j$ in population and find the cosine between them using the following formula:

\begin{gather}
    \mathbf{s}^{i}(j) = \dfrac{\mathbf{r}^i(\mathbf{r}^j)^{\top}}{\left \| \mathbf{r}^{i} \right \| \times \left \| \mathbf{r}^{j} \right \|}; \forall \:\: 1 \leq j \leq n_s
\end{gather}

The intuition behind this step is that if two embeddings are similar at a particular timepoint, they deviate from the CBT in the same way thus their residuals will also be similar. Notably, if the angle between two residual vectors is smaller then the cosine value will be higher. Next, we select top $K$ subjects with the highest cosine similarities with the testing subject $i$. Finally, we average the brain networks of the $K$ selected subjects at follow-up timepoints to predict the evolution trajectory of the testing network $\mathbf{\hat{X}}^i_{t_h}$ with $h \in \{1, \dots, n_{t} \}$.

\begin{figure}[h]
\centering
\includegraphics[width=12cm]{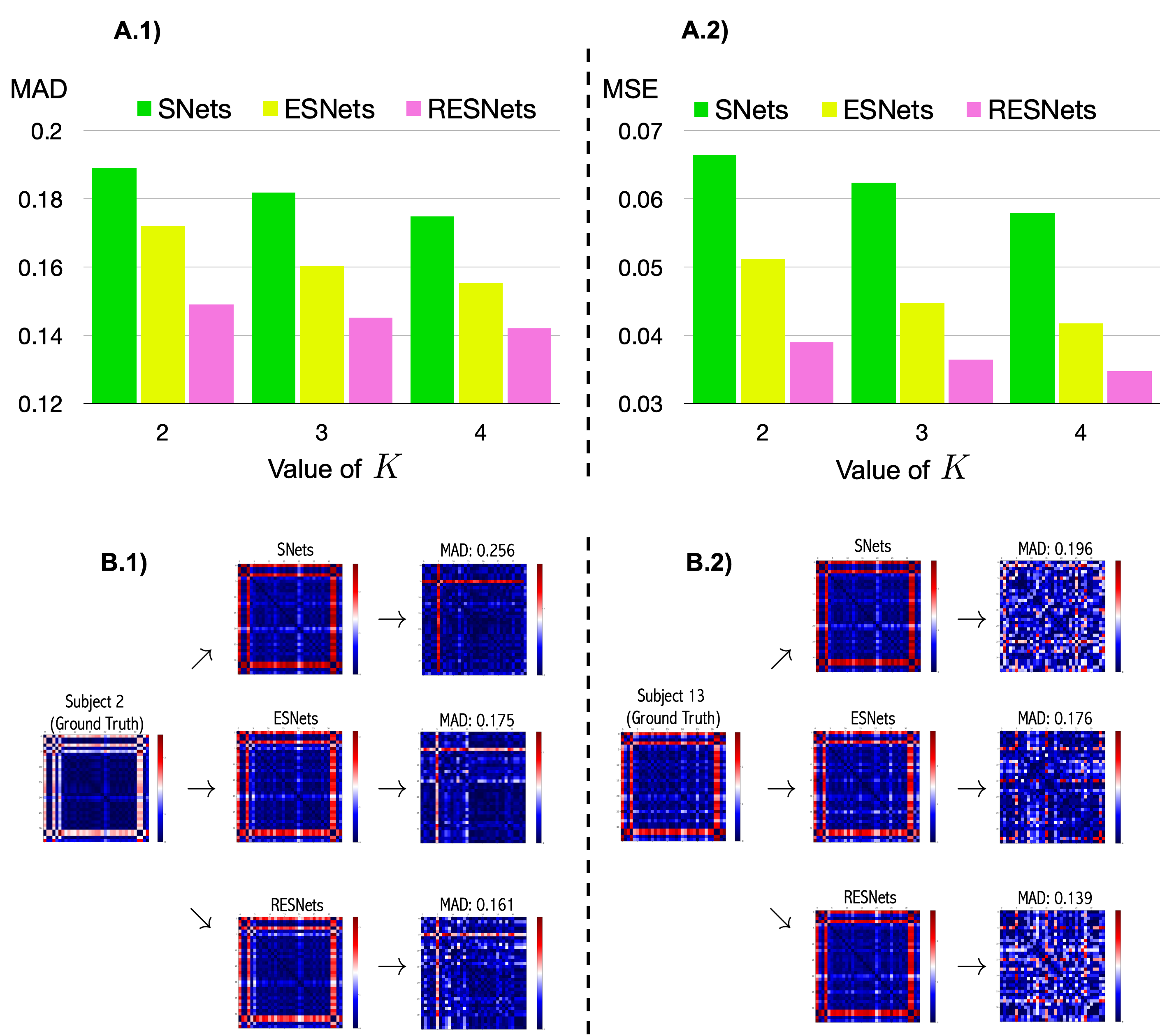}
\caption{ \textbf{A.1-2)}\emph{Comparison of RESNets against baseline methods using Mean Absolute Deviance (MAD) and Mean Squared Error (MSE).} SNets: Similarity-based Network selection method. ESNets: Embedding Similarity-based Network selection method. RESNets: the proposed Residual Embedding Similarity-based Network selection method. \textbf{B.1-2)} We display the predicted brain networks at a 6-month follow-up timepoint by RESNets and its ablated versions for two random subjects.}
\label{results_figure}
\end{figure}

\section{Results and Discussion}

\textbf{Evaluation dataset.} We evaluated our framework on 67 subjects (35 diagnosed with Alzheimer's disease and 32 diagnosed with late mild cognitive impairment) from ADNI GO public dataset\footnote{\url{http://adni.loni.usc.edu}} using leave-one-out cross validation. Each subject has two structural T1-w MR images 
acquired at baseline and 6-months follow-up. We used FreeSurfer to reconstruct both right and left cortical hemispheres for each subject from T1-w MRI. 
Next, we parcellated each cortical hemisphere into 35 cortical ROIs using Desikan-Killiany Atlas. For each subject, we constructed morphological brain networks (MBN) at each timepoint using the method introduced in \cite{Mahjoub:2018}. 
We used the mean cortical thickness measure where the morphological connectivity strength between two regions is defined as the absolute difference between the average cortical thickness in each ROI. 

\textbf{Parameter setting.} Our encoder comprises three hidden layers of 16 neurons. The second hidden layer of the encoder is a Gaussian noise layer with $\sigma = 0.1$. We construct the discriminator with 64- and 16-neuron hidden layers. Both encoder and discriminator learning rates and number of iterations are set to 0.005 and 30, respectively. For the brain network prediction step, we vary the number of selected neighbors $K$ between $2$ and $4$ and report results in \textbf{Fig.}~\ref{results_figure}--A.

\textbf{Comparison methods and evaluation.} We compare the performance of our RESNets framework with two baseline methods: \textbf{(1) Similarity-based Network selection (SNets):} is a variant of our framework where the similarities are defined as the dot product between raw feature vectors of brain networks without any embedding. Note that such strategy is adopted in the state-of-the-art work \cite{baha} \textbf{(2) Embedding Similarity-based Network selection (ESNets):} is an ablated version of RESNets where the similarities are computed as the dot product between learned \emph{embedded} brain networks.  \textbf{Fig.}~\ref{results_figure}-A.1 and \textbf{Fig.}~\ref{results_figure}-A.2  shows the Mean Absolute Deviance (MAD) and Mean Squared Error (MSE) between the ground truth and predicted testing networks at follow-up timepoint, respectively. Clearly, our RESNets framework consistently achieves the best prediction performance using different $K$ selected neighbors to the testing subject. This demonstrates that our proposed similarity metric using the CBT-based residual embeddings boosts the network evolution prediction accuracy. We display in \textbf{Fig.}~\ref{results_figure}-B.1 and \textbf{Fig.}~\ref{results_figure}-B.2 the residual prediction error computed using Mean Absolute Deviance (MAD) between the ground truth and predicted follow-up brain network for two representative subjects. This clearly shows that our framework leads to a low network residual prediction error in comparison to its variants. There are several exciting research directions to take this work further such as designing a joint end-to-end sample selection and prediction framework within a unified geometric deep learning framework.

\section{Conclusion}

We proposed a novel brain network evolution trajectory prediction framework from a single timepoint rooted in (i) learning adversarial topology-preserving embeddings of brain networks and (ii) a normalization step with respect to population center, namely connectional brain template, allowing us to compute residual similarities between a testing subject and training subjects for effective sample selection. Our RESNets framework can better identify the most similar training brain networks to a given testing network at baseline. Furthermore, RESNets  outperformed baseline methods on autism spectrum disorder connectomic dataset. In our future work, we plan to generalize RESNets to handle multi-view brain network evolution prediction where different follow-up views are simultaneously predicted. Specifically, we aim to learn a joint embedding of all network views using ACE encoder and leverage the generalized netNorm for multi-view CBT estimation \cite{cbt} to produce a population template that integrates different brain views.

\section{Supplementary material}

We provide three supplementary items on RESNets for reproducible and open science:

\begin{enumerate}
	\item A 6-mn YouTube video explaining how RESNets works on BASIRA YouTube channel at \url{https://youtu.be/UOUHe-1FfeY}.
	\item RESNets code in Python on GitHub at \url{https://github.com/basiralab/RESNets}. 
	\item A GitHub video code demo on BASIRA YouTube channel at \url{https://youtu.be/R0zdqTwGf_c}. 
\end{enumerate}

\section{Acknowledgement}

I. Rekik is supported by the European Union's Horizon 2020 research and innovation programme under the Marie Sklodowska-Curie Individual Fellowship grant agreement No 101003403 (\url{http://basira-lab.com/normnets/}).

\bibliography{Biblio3}

\begin{thebibliography}{10}

\bibitem{Yang:2019}
Yang, Q., Thomopoulos, S.I., Ding, L., Surento, W., Thompson, P.M., Jahanshad,
  N., Initiative, A.D.N.,  et~al.:
\newblock Support vector based autoregressive mixed models of longitudinal
  brain changes and corresponding genetics in alzheimer's disease.
\newblock International Workshop on PRedictive Intelligence In MEdicine (2019)
  160--167

\bibitem{Zhou:2019}
Zhou, Y., Tagare, H.D.:
\newblock Bayesian longitudinal modeling of early stage parkinson's disease
  using datscan images.
\newblock International Conference on Information Processing in Medical Imaging
  (2019)  405--416

\bibitem{Rekik:2017a}
Rekik, I., Li, G., Lin, W., Shen, D.:
\newblock Estimation of brain network atlases using diffusive-shrinking graphs:
  application to developing brains.
\newblock International Conference on Information Processing in Medical Imaging
  (2017)  385--397

\bibitem{Gafuroglu:2018}
Gafuro{\u{g}}lu, C., Rekik, I.,  et~al.:
\newblock Joint prediction and classification of brain image evolution
  trajectories from baseline brain image with application to early dementia.
\newblock International Conference on Medical Image Computing and
  Computer-Assisted Intervention (2018)  437--445

\bibitem{consistent_aging}
Xia, T., Chartsias, A., Tsaftaris, S.A., Initiative, A.D.N.,  et~al.:
\newblock Consistent brain ageing synthesis.
\newblock International Conference on Medical Image Computing and
  Computer-Assisted Intervention (2019)  750--758

\bibitem{baha}
Ezzine, B.E., Rekik, I.:
\newblock Learning-guided infinite network atlas selection for predicting
  longitudinal brain network evolution from a single observation.
\newblock International Conference on Medical Image Computing and
  Computer-Assisted Intervention (2019)  796--805

\bibitem{Fornito:2015}
Fornito, A., Zalesky, A., Breakspear, M.:
\newblock The connectomics of brain disorders.
\newblock Nature Reviews Neuroscience \textbf{16} (2015)  159--172

\bibitem{Wang:2020}
Wang, J., Zhang, L., Wang, Q., Chen, L., Shi, J., Chen, X., Li, Z., Shen, D.:
\newblock Multi-class {ASD} classification based on functional connectivity and
  functional correlation tensor via multi-source domain adaptation and
  multi-view sparse representation.
\newblock IEEE Transactions on Medical Imaging (2020)

\bibitem{Richiardi:2010}
Richiardi, J., Van De~Ville, D., Riesen, K., Bunke, H.:
\newblock Vector space embedding of undirected graphs with fixed-cardinality
  vertex sequences for classification.
\newblock 2010 20th International Conference on Pattern Recognition (2010)
  902--905

\bibitem{Bassett:2017}
Bassett, D.S., Sporns, O.:
\newblock Network neuroscience.
\newblock Nature neuroscience \textbf{20} (2017)  353

\bibitem{cbt}
Dhifallah, S., Rekik, I.:
\newblock Estimation of connectional brain templates using selective multi-view
  network normalization.
\newblock Medical Image Analysis \textbf{59} (2019)  101567

\bibitem{Liu:2015}
Liu, M., Zhang, D., Shen, D., Initiative, A.D.N.:
\newblock View-centralized multi-atlas classification for alzheimer's disease
  diagnosis.
\newblock Human brain mapping \textbf{36} (2015)  1847--1865

\bibitem{ace}
Banka, A., Rekik, I.:
\newblock Adversarial connectome embedding for mild cognitive impairment
  identification using cortical morphological networks.
\newblock International Workshop on Connectomics in Neuroimaging (2019)  74--82

\bibitem{Bronstein:2017}
Bronstein, M.M., Bruna, J., LeCun, Y., Szlam, A., Vandergheynst, P.:
\newblock Geometric deep learning: going beyond euclidean data.
\newblock IEEE Signal Processing Magazine \textbf{34} (2017)  18--42

\bibitem{Hamilton:2017}
Hamilton, W.L., Ying, R., Leskovec, J.:
\newblock Representation learning on graphs: Methods and applications.
\newblock arXiv preprint arXiv:1709.05584 (2017)

\bibitem{Kipf:2016}
Kipf, T.N., Welling, M.:
\newblock Semi-supervised classification with graph convolutional networks.
\newblock arXiv preprint arXiv:1609.02907 (2016)

\bibitem{Wang:2014}
Wang, B., Mezlini, A., Demir, F., Fiume, M., \emph{et al.}:
\newblock Similarity network fusion for aggregating data types on a genomic
  scale.
\newblock Nat Methods \textbf{11} (2014)  333--337

\bibitem{Mahjoub:2018}
Mahjoub, I., Mahjoub, M.A., Rekik, I.:
\newblock Brain multiplexes reveal morphological connectional biomarkers
  fingerprinting late brain dementia states.
\newblock Scientific reports \textbf{8} (2018)  1--14

\end{thebibliography}
\bibliographystyle{splncs}
\end{document}